\documentclass[a4paper, 11pt]{article}

\usepackage[utf8]{inputenc}
\usepackage[T1]{fontenc}
\usepackage{amsmath,amssymb,amsfonts}
\usepackage{mathrsfs}
\usepackage{mathtools}
\usepackage{float}
\usepackage{algorithm}
\usepackage{algorithmic}
\usepackage{geometry}
\usepackage{url}
\usepackage{cite}
\title{\Large \textbf{Localized Latent Editing for Dose-Response Modeling in\\Botulinum Toxin Injection Planning}}

\author{%
Estèphe Arnaud$^{1,4}$, Mohamed Daoudi$^{1,2}$, and Pierre Guerreschi$^3$\\
\small
$^1$ Univ. Lille, CNRS, Centrale Lille, Institut Mines-Télécom, UMR 9189 CRIStAL, F-59000 Lille, France\\
\small
$^2$ IMT Nord Europe, Institut Mines-Télécom, Univ. Lille, Centre for Digital Systems, F-59000 Lille, France\\
\small
$^3$ Lille University Hospital (CHU), Department of Plastic Surgery, F-59000 Lille, France\\
\small
$^4$ SATT Nord, F-59000 Lille, France\\
\small
}

\date{}

\begin{document}

\maketitle

%%%%%%%%%%%%%%%%%%%%%%%%%%%%%%%%%%%%%%%%%%%%%%%%%%%%%%%%%%%%%%%%%%%%%%%%%%%%%%%%
\begin{abstract}
Botulinum toxin (Botox) injections are the gold standard for managing facial asymmetry and aesthetic rejuvenation, yet determining the optimal dosage remains largely intuitive, often leading to suboptimal outcomes. We propose a localized latent editing framework that simulates Botulinum Toxin injection effects for injection planning through dose-response modeling. Our key contribution is a \textit{Region-Specific Latent Axis Discovery} method that learns localized muscle relaxation trajectories in StyleGAN2's latent space, enabling precise control over specific facial regions without global side effects. By correlating these localized latent trajectories with injected toxin units, we learn a predictive dose-response model. We rigorously compare two approaches: direct metric regression versus image-based generative simulation on a clinical dataset of $N=360$ images from 46 patients. On a hold-out test set, our framework demonstrates moderate-to-strong structural correlations for geometric asymmetry metrics, confirming that the generative model correctly captures the direction of morphological changes. While biological variability limits absolute precision, we introduce a hybrid "Human-in-the-Loop" workflow where clinicians interactively refine simulations, bridging the gap between pathological reconstruction and cosmetic planning.
\end{abstract}

%%%%%%%%%%%%%%%%%%%%%%%%%%%%%%%%%%%%%%%%%%%%%%%%%%%%%%%%%%%%%%%%%%%%%%%%%%%%%%%%
\section{INTRODUCTION}

Facial aesthetics and symmetry depend on the delicate balance of mimetic muscles. In pathologies like facial paralysis \cite{szczapa2019}, or in aesthetic medicine (anti-aging), Botulinum toxin type A (BoNT-A) is used to perform "chemodenervation", selectively weakening specific muscles to restore symmetry or smooth wrinkles. Despite its popularity, BoNT-A injection planning remains uncertain. Toxin diffusion, muscle depth, and individual metabolic responses create a complex "black box" between the \textit{input} (dose in Units) and the \textit{output} (morphological change). No computational tool currently exists to visualize the specific effect of a 5-Unit injection in the \textit{depressor anguli oris} versus a 2-Unit injection. Predicting and visualizing treatment outcomes before injection would significantly improve clinical decision-making and patient counseling.

StyleGAN2 \cite{karras2020} offers a potential solution through high-fidelity simulation. However, applying GANs faces the "Entanglement Challenge": increasing a "smile" vector in latent space typically changes the entire face, whereas a clinical injection is highly localized. The "Calibration Challenge" arises from the non-deterministic nature of biological tissues, making simple regression models insufficient on small datasets. By leveraging StyleGAN2's pre-trained latent space $W+$, which already encodes rich morphological information from large-scale facial datasets, we can learn Botox-specific transformation axes with relatively few clinical examples, as the latent space's structural priors enable efficient learning of localized dose-response relationships that would require significantly more data in raw pixel space.

In this work, we present a unified localized latent editing framework for dose-response modeling. Our contributions are:
1.  \textbf{Localized Axis Discovery:} A robust "Patch-and-Encode" method that isolates the latent direction of specific muscle groups (e.g., peri-oral, peri-orbital) without side effects on the rest of the face.
2.  \textbf{Comparative Predictive Modeling:} We rigorously compare two predictive paradigms: predicting the visual outcome first (Image Prediction) vs predicting the measurements directly (Metric Prediction).
3.  \textbf{Latent Dose-Response Model:} A formulation linking the latent intensity vector $\boldsymbol{\alpha}$ in latent space to clinical units ($\mathbf{u}$), enabling meaningful simulation.

%%%%%%%%%%%%%%%%%%%%%%%%%%%%%%%%%%%%%%%%%%%%%%%%%%%%%%%%%%%%%%%%%%%%%%%%%%%%%%%%
\section{RELATED WORK}

\subsection{Botulinum Toxin Planning}
Current clinical planning relies on static 2D photography and subjective scales \cite{szczapa2019}, providing only qualitative assessments that cannot predict visual outcomes for specific dose configurations. Finite Element Models (FEM) \cite{terzopoulos1990} attempt biomechanical simulation but suffer from limited realism and require extensive parameterization difficult to calibrate from clinical data. In contrast, our data-driven approach learns quantitative dose-response relationships from pre/post-treatment image pairs, enabling predictive visualization before treatment without explicit biomechanical modeling.

\subsection{Generative Facial Editing}
While StyleGAN inversion \cite{alaluf2022} enables high-fidelity facial editing, most latent directions are semantic (e.g., "Gender", "Age") and not directly aligned with anatomical actions. Global approaches like InterfaceGAN \cite{shen2020} tend to suffer from entanglement, while StyleRig \cite{stylerig} relies on a 3D Morphable Model (3DMM) \cite{blanz1999} whose mesh limits the range of expressible deformations. Interactive methods such as DragGAN \cite{pan2023draggan} allow point-based manipulation on the image manifold but remain computationally demanding and frequently induce globally coupled deformations (e.g., moving the right eyebrow also affects the left), which is misaligned with the localized nature of Botulinum toxin injections.

Diffusion Models \cite{rombach2022} have recently shown impressive performance in high-resolution image synthesis, and several methods enable local editing through inpainting or masked conditioning \cite{avrahami2022blended,simsar2023lime,kouzelis2024local,corneanu2024latentpaint}. However, for our dose-response modeling application, StyleGAN2's $W+$ space offers critical advantages: (i) \textbf{Interpretable Latent Axes:} Our method identifies patient-specific axes $\mathbf{v}_k$ encoding progressive muscle relaxation directions, enabling parametric control via $\boldsymbol{\alpha}$ for dose-response regression through the linear structure $w_{final} = w_{src} + \sum \alpha_k \mathbf{v}_k$. Local editing in diffusion models operates through iterative denoising but does not yield such stable, interpretable axes suitable for regression. (ii) \textbf{Manifold Regularity \& Interactivity:} Linear combinations in $W+$ produce smooth, predictable morphological transitions crucial for correlating dose units $\mathbf{u}$ with visual outcomes, whereas diffusion models' noise-based representation lacks this linear structure. Additionally, our human-in-the-loop workflow requires sub-second generation for interactive $\boldsymbol{\alpha}$ adjustment, whereas diffusion models require 10-20 denoising steps even for local edits. We therefore favor StyleGAN2, whose $W+$ manifold is optimized for facial topology and offers a regular latent surface where linear interpolations correspond to continuous morphological changes, enabling our dose-response framework.

%%%%%%%%%%%%%%%%%%%%%%%%%%%%%%%%%%%%%%%%%%%%%%%%%%%%%%%%%%%%%%%%%%%%%%%%%%%%%%%%
\section{METHODOLOGY}

\subsection{Data Acquisition Protocol}
We compiled a dataset of $N=360$ images from 46 patients undergoing BoNT-A therapy at the Lille University Hospital. The acquisition protocol was standardized to minimize environmental noise: patients were photographed at a fixed distance of 1 meter using a high-resolution sensor (iPhone 12 Pro), under uniform lighting, against a neutral background. To capture the full dynamic range of facial muscles, each patient performed a sequence of 8 standardized expressions: neutral, brow raise, gentle eye closure, tight eye closure, symmetric smile, strong smile (with/without teeth), lip pucker, and a phonation task. The dataset was split by patient identity (80/20 ratio) into a training set ($N=289$ images from 37 patients) and a hold-out test set ($N=71$ images from 9 patients), ensuring that no patient from the training set appears in the evaluation.

\subsection{Data Preprocessing and Alignment}
Strict alignment is crucial. We adopt the standard preprocessing pipeline from the FFHQ dataset \cite{karras2019stylegan}. For every patient image $I$, we compute the similarity transform $T$ (rotation, scale, translation) that minimizes the $L_2$ distance between the patient's eyes/mouth landmarks and the canonical template, ensuring a horizontal eye alignment.

\subsection{Latent Space Inversion}
We project the aligned patient image $x$ ($256 \times 256$) into the extended latent space $W+ \subset \mathbb{R}^{18 \times 512}$ using a HyperStyle-based encoder $E$ \cite{alaluf2022}, obtaining a latent code $w = E(x)$ that compactly represents the patient. Since pathological faces with severe asymmetry fall out-of-distribution (OOD) relative to the encoder's training data (mostly healthy subjects), a single forward pass yields suboptimal reconstructions. Following standard StyleGAN inversion practices \cite{alaluf2022}, we therefore perform a short test-time refinement (30 gradient steps) on $w$ to minimize a reconstruction loss combining pixel-wise $L_2$, perceptual LPIPS and identity preservation.

To better preserve clinically relevant structures, we up-weight $L_2$ errors on eyes and mouth crops, while a mild regularization term keeps $w$ close to the mean latent code $\bar{w}$. This region-weighted inversion yields a source latent code $w_{src}$ that faithfully preserves the patient's identity while offering precise geometric alignment in the peri-orbital and peri-oral areas that are critical for chemodenervation planning.

\subsection{Region-Specific Latent Axis Discovery}
Unlike previous methods that learn a single global direction, we decompose the face into $K=6$ anatomical Regions of Interest (ROIs): Left/Right Eyebrow, Left/Right Eye, and Left/Right Mouth.

\subsubsection{Automatic ROI Definition}
For each anatomical region $k$ (left/right eyebrow, left/right eye, left/right mouth corner), we define a fixed rectangular Region of Interest (ROI) in the aligned face image. These ROIs were calibrated once on frontal FFHQ-like faces \cite{karras2019stylegan} so as to (i) fully enclose the target structure for all subjects, and (ii) avoid overlap between regions (e.g., the eyebrow box never includes eye pixels).

Each ROI $B_k$ is implemented as a simple binary mask $M_k$ that takes value 1 inside the rectangle and 0 outside. This design fixes the spatial support of each latent axis and guarantees that, on aligned frontal faces, each ROI consistently captures the same anatomical zone without contaminating neighboring regions.

\subsubsection{Algorithm and Manifold Projection}
We employ a "Naive Patching + Manifold Projection" strategy to isolate the latent axes. For each patient, we generate a synthetic symmetric target $I_{tgt}$ without requiring a true post-operative image: we mirror the left and right halves respectively, encode them to obtain $w_{left}$ and $w_{right}$, then compute the midpoint $w_{sym} = (w_{left} + w_{right})/2$ in latent space. Decoding yields $I_{tgt} = G(w_{sym})$, a realistic symmetric target leveraging StyleGAN2's latent structure where linear interpolations correspond to smooth morphological transitions. The process is formalized in Algorithm 1.

\begin{algorithm}
\caption{Localized Axis Discovery}
\begin{algorithmic}[1]
\REQUIRE Source Image $I_{src}$, Target Reference $I_{tgt}$
\REQUIRE Binary ROI Masks $\{M_1, ..., M_K\}$ defined above
\STATE Encode source: $w_{src} \leftarrow E(I_{src})$
\FOR{$k=1$ to $K$}
    \STATE \textit{// Naive Semantic Patching}
    \STATE $I_{hybrid,k} \leftarrow I_{src} \cdot (1 - M_k) + I_{tgt} \cdot M_k$
    \STATE \textit{// Manifold Projection: re-encode to heal seams}
    \STATE $w_{hybrid,k} \leftarrow E(I_{hybrid,k}) \in W+$
    \STATE \textit{// Extract direction vector}
    \STATE $\mathbf{v}_k \leftarrow w_{hybrid,k} - w_{src} \in \mathbb{R}^{18 \times 512}$
\ENDFOR
\RETURN Basis Vectors $\{\mathbf{v}_1, ..., \mathbf{v}_K\}$
\end{algorithmic}
\end{algorithm}

\textit{Manifold Projection as a Natural Blender:}
In Step 4 of Algorithm 1, the naive "cut-and-paste" operation introduces visible seams. However, re-encoding the hybrid image projects it onto the $W+$ manifold, finding the closest realistic face representation in latent space. Since $W+$ cannot represent sharp artificial seams, the reconstructed image $G(w_{hybrid,k})$ is automatically "healed" with coherent skin texture across boundaries. This property enables photorealistic, seamless edits with simple binary masks. The vector $\mathbf{v}_k$ captures the \textit{morphological essence} of the change (e.g., lifting the mouth corner) stripped of artifacts.

\subsection{Quantitative Facial Metrics}
We evaluate our framework using two families of geometric asymmetry metrics derived from 468 MediaPipe landmarks \cite{lugaresi2019}, normalized by inter-pupillary distance.

\textbf{Morphological Asymmetry (Procrustes Analysis):} We compute Procrustes distances between mirrored left/right landmark configurations, filtering out translation and rotation. This yields: (i) \textit{Eyebrows Asymmetry} (divergence between eyebrow arches), (ii) \textit{Eyes Asymmetry} (peri-orbital structural differences), (iii) \textit{Furrow (Nasolabial)} (asymmetry of the groove between nose and mouth corners, using landmarks 202, 212, 216, 206, 203, 129, 209, 126 and counterparts), and (iv) a \textit{Total Asymmetry Score} aggregating Procrustes distances across all regions.

\textbf{Positional Asymmetry (Ratio Analysis):} To quantify positional asymmetries, we apply the symmetry ratio $S = 1 - \min(L/R, R/L)$ where $L$ and $R$ are left/right measurements. \textit{Outer Eyebrow-to-Nose} compares distances from outer eyebrow corners to nose tip, detecting vertical height discrepancies, while \textit{Mouth Angle} measures deviation from a horizontal axis (90° ideal), detecting lip tilting.

\subsection{Dose-Response Simulation Model}
The final simulated latent code $w_{final}$ is a linear combination of local displacements:
\begin{equation}
w_{final} = w_{src} + \sum_{k=1}^{K} \alpha_k \cdot \mathbf{v}_k,
\end{equation}

where $\alpha_k$ controls how closely the generated face approaches the target symmetric state in region $k$. The latent response is defined on $K=6$ regions, but clinically, injections target $J=22$ specific muscles. We model the forward mapping (Dose $\to$ Visual) from $\mathbf{u} \in \mathbb{R}^{J}$ to $\boldsymbol{\alpha} \in \mathbb{R}^{K}$ using a Gradient Boosting Regressor, while the inverse mapping (Visual $\to$ Dose) is estimated via numerical optimization for interactive workflows. We investigate two strategies to predict the outcome of a treatment defined by a dose vector $\mathbf{u}$, reporting results for the 6 geometric asymmetry metrics $\mathbf{m} \in \mathbb{R}^{6}$ described above.

\subsubsection{Approach A: Image Prediction (Generative)}
This approach predicts the visual outcome by first generating an image, then extracting metrics. This generative pipeline enables visual simulation for clinical planning while ensuring anatomical plausibility through StyleGAN2's learned manifold. The workflow requires patient-specific axes $\{\mathbf{v}_1, ..., \mathbf{v}_K\}$ computed via Algorithm 1 for each patient.

\textbf{Ground Truth Construction:} Since latent intensities $\boldsymbol{\alpha}$ are not directly observable, we compute ground-truth values $\boldsymbol{\alpha}_{GT}$ via Analysis-by-Synthesis for each training patient with dose $\mathbf{u}$ and post-operative image $I_{post}$. We optimize $\boldsymbol{\alpha}$ (initialized at $\mathbf{0}$) such that the generated image $G(w_{src} + \sum_{k=1}^{K} \alpha_k \cdot \mathbf{v}_k)$ minimizes the metric distance to $I_{post}$. This optimization uses a differentiable face alignment network \cite{bulat2017} to backpropagate errors from predicted metrics $\mathcal{E}(G(w_{src} + \sum_{k=1}^{K} \alpha_k \cdot \mathbf{v}_k))$ to ground-truth metrics $\mathcal{E}(I_{post})$ directly into the latent parameters $\boldsymbol{\alpha}$.

\textbf{Learning and Inference:} We train a Gradient Boosting Regressor $f_{gen}: (\mathbf{u}, \mathbf{m}_{src}) \in \mathbb{R}^{J+6} \to \boldsymbol{\alpha} \in \mathbb{R}^{K}$ to map dose and pre-operative metrics to latent intensities, minimizing the $L_2$ distance between predicted $\boldsymbol{\alpha}$ and ground-truth $\boldsymbol{\alpha}_{GT}$ across the training set. Including $\mathbf{m}_{src}$ as input is crucial because patients with different baseline asymmetry levels exhibit distinct dose-response relationships, enabling the model to calibrate predictions according to each patient's asymmetry profile. For inference on a new patient with pre-operative latent code $w_{src}$, metrics $\mathbf{m}_{src}$, and dose $\mathbf{u}$, we: (i) compute patient-specific axes $\{\mathbf{v}_k\}$ via Algorithm 1, (ii) predict $\boldsymbol{\alpha} = f_{gen}(\mathbf{u}, \mathbf{m}_{src})$, (iii) compute $\Delta w = \sum_{k=1}^{K} \alpha_k \cdot \mathbf{v}_k$, (iv) generate $\hat{I}_{post} = G(w_{src} + \Delta w)$, and (v) extract metrics $\mathbf{\hat{m}}_{post} = \mathcal{E}(\hat{I}_{post})$.

\subsubsection{Approach B: Metric Prediction (Direct)}
We train a Gradient Boosting Regressor $f_{reg}$ to predict the relative change in metrics directly from the dose and initial state: $\Delta \mathbf{m} = f_{reg}(\mathbf{u}, \mathbf{m}_{src})$. This approach bypasses the image generation bottleneck, enabling more efficient and precise predictions for well-defined geometric relationships. By directly modeling metric variations, we capture non-linear relationships and interactions between the initial geometry and treatment dose without the additional complexity of image synthesis. The predicted post-operative state is reconstructed as $\mathbf{\hat{m}}_{post} = \mathbf{m}_{src} \cdot (1 + \Delta \mathbf{m})$.

%%%%%%%%%%%%%%%%%%%%%%%%%%%%%%%%%%%%%%%%%%%%%%%%%%%%%%%%%%%%%%%%%%%%%%%%%%%%%%%%
\section{RESULTS AND DISCUSSION}

\subsection{Experimental Setup and Evaluation}
Models were trained on the training split ($N=289$) and evaluated on the test set ($N=71$), using post-treatment clinical images and their associated metrics as ground truth. We utilized the full clinical cohort without excluding lower-quality samples to test robustness to real-world acquisition noise. We evaluate both approaches on geometric asymmetry metrics that directly measure the therapeutic effect of Botulinum toxin injections. Predictions are evaluated on relative variation ($\Delta \mathbf{m}$) rather than absolute values, as absolute post-operative values would artificially inflate performance metrics due to the strong correlation between pre- and post-operative states, whereas $\Delta \mathbf{m}$ isolates the model's ability to capture the specific therapeutic effect.

\subsection{Analysis of Results}

Table~\ref{tab:fused_results} reveals that the generative pipeline (Approach A) achieves its best performance on \textit{Eyebrows Asymmetry} ($R^2=0.34, r=0.61$), where the $W+$ structural prior helps maintain anatomically plausible brow reshaping. However, the direct regressor (Approach B) consistently outperforms A on distance-based metrics: \textit{Eyes Asymmetry} ($R^2=0.42$ vs.\ 0.21, $r=0.68$ vs.\ 0.50), \textit{Outer Eyebrow-to-Nose} ($R^2=0.67$ vs.\ 0.28, $r=0.82$ vs.\ 0.68), and \textit{Mouth Angle} ($R^2=0.14$ vs.\ 0.00). This indicates that when geometric relationships are well-defined and directly impacted by injection patterns, metric-based prediction is more efficient than passing through an image synthesis bottleneck. Despite being trained purely in the image domain, Approach A yields metric predictions that remain quantitatively consistent with the direct regressor. For several metrics (e.g., \textit{Eyebrows Asymmetry}, \textit{Total Asymmetry}), Approach A attains comparable MAE and moderate correlations, suggesting that enforcing anatomical plausibility in $W+$ is sufficient for downstream geometric estimators to extract meaningful trends, ensuring clinical interpretability.

A key finding is the discrepancy between $R^2$ and correlation coefficients ($r$). While $R^2$ values are modest, reflecting the difficulty of predicting the exact magnitude ($\Delta \mathbf{m}$) of a biological response on a small, noisy clinical dataset, the consistently moderate-to-strong Pearson correlations ($r \approx 0.40$--$0.82$) indicate that both models reliably capture the \textbf{direction and relative scale} of the therapeutic effect. For a decision-support tool, predicting the "asymmetry reduction trend" is often as critical as absolute numerical precision. In practice, we recommend Approach B when the primary goal is to obtain quantitative estimates on specific geometric metrics, and Approach A when visually coherent, anatomically plausible simulations are needed to support human-in-the-loop planning.

\begin{table}[t]
\centering
\caption{Performance Comparison: Approach A (Generative) vs. Approach B (Direct).}
\label{tab:fused_results}
\scriptsize
\renewcommand{\arraystretch}{1.0}
\begin{tabular}{|l|c|c|c|c|c|c|}
\hline
\textbf{Metric ($\Delta m$)} & \multicolumn{2}{c|}{\textbf{MAE}} & \multicolumn{2}{c|}{\textbf{$R^2$}} & \multicolumn{2}{c|}{\textbf{Pearson R}} \\ \cline{2-7} 
 & \textbf{A} & \textbf{B} & \textbf{A} & \textbf{B} & \textbf{A} & \textbf{B} \\ \hline
Eyebrows Asym. & 1.06 & 1.10 & \textbf{0.34} & 0.19 & \textbf{0.61} & 0.50 \\ \hline
Eyes Asym. & 0.95 & 0.81 & 0.21 & \textbf{0.42} & 0.50 & \textbf{0.68} \\ \hline
Furrow & 1.64 & 2.09 & \textbf{0.09} & -0.06 & 0.38 & \textbf{0.47} \\ \hline
Outer Eyebr.-Nose & 2.39 & 2.10 & 0.28 & \textbf{0.67} & 0.68 & \textbf{0.82} \\ \hline
Mouth Angle & 0.73 & 0.77 & 0.00 & \textbf{0.14} & 0.25 & \textbf{0.40} \\ \hline
Total Asym. & 0.60 & 0.60 & \textbf{-0.02} & -0.12 & \textbf{0.45} & 0.42 \\ \hline
\end{tabular}
\end{table} 

\subsection{Limitations and Future Work}
Our framework faces several inherent limitations. The dataset size ($N=360$ images from 46 patients) constrains the model's ability to capture the full spectrum of biological variability. The moderate prediction performance reflects the difficulty of modeling non-deterministic biological processes, where unobserved factors (e.g., individual metabolism, injection depth, muscle fiber composition) introduce aleatoric uncertainty. Moreover, our evaluation is limited to a single clinical center, and generalization to other populations or acquisition protocols remains to be assessed. These limitations do not diminish the framework's value as a simulation tool. The moderate correlations confirm that the generative model correctly captures the \textit{direction} of morphological changes, which is sufficient for interactive clinical planning where the clinician retains final control. Future work will focus on: (i) expanding the dataset through multi-center collaborations, (ii) incorporating patient-specific factors (age, skin type, muscle mass) into the dose-response model, and (iii) validating clinical utility through prospective studies with practicing clinicians.

\subsection{Proposed Workflow: Human-in-the-Loop Simulation}
Given these findings, we pivot from a "Black Box Predictor" to an "Assistive Simulation" workflow.
\begin{enumerate}
    \item \textbf{Initialization:} The AI generates a simulation using the learned mean response (Approach A).
    \item \textbf{Refinement:} The clinician adjusts the simulation intensity visually using a slider, effectively modifying the latent intensities $\boldsymbol{\alpha}$. The system inversely maps this new $\boldsymbol{\alpha}$ back to clinical units, displaying the updated dose $\mathbf{u}_{new}$ required to achieve the desired morphology.
    \item \textbf{Feedback Loop:} The validated pair ($\mathbf{u}_{new}$, Morphological Outcome) is recorded to refine the calibration function $\boldsymbol{\alpha} = f(\mathbf{u})$ for future patients.
\end{enumerate}

\subsection{Extension to Cosmetic Medicine}
The framework's modularity enables a seamless transfer from reconstructive symmetry to aesthetic planning by redefining the optimization target. By mapping clinical doses to semantic trajectories in $W+$, we enable: (i) \textbf{Quantified Rejuvenation:} We derive a rejuvenation axis $\mathbf{v}_{age}$ by identifying the linear semantic direction separating age-related latent states in $W+$ \cite{shen2020}, where BoNT-A dose $\mathbf{u}$ acts as a scalar controller allowing clinicians to calibrate units against a desired age-delta. (ii) \textbf{Aesthetic Volumization:} Using ROI masks $M_k$, we isolate filler-specific axes through latent discovery and repurpose Analysis-by-Synthesis to estimate $\boldsymbol{\alpha}$ values required to reach target \textit{Attractiveness} \cite{liang2018} or \textit{Facial Harmony} \cite{prendergast2012} scores, correlating them with injected volumes to map precise doses to predicted aesthetic trajectories.

%%%%%%%%%%%%%%%%%%%%%%%%%%%%%%%%%%%%%%%%%%%%%%%%%%%%%%%%%%%%%%%%%%%%%%%%%%%%%%%%

\section{CONCLUSIONS}

We presented a localized latent editing framework for Botulinum Toxin injection planning through dose-response modeling, capable of predicting localized chemodenervation effects. By rigorously comparing Image-based vs Metric-based prediction on a clinical cohort of $N=360$ images, we demonstrated that moderate structural correlations for geometric features confirm our generative model captures the correct direction of morphological changes, making it a valuable tool for interactive clinical planning despite biological variability and the challenges of dose-response modeling on small datasets.

%%%%%%%%%%%%%%%%%%%%%%%%%%%%%%%%%%%%%%%%%%%%%%%%%%%%%%%%%%%%%%%%%%%%%%%%%%%%%%%%
\section{ACKNOWLEDGMENTS}

FaceToFace has been funded by la Région-Hauts-de-France  under Start-AiRR program and by SATT Nord.

%%%%%%%%%%%%%%%%%%%%%%%%%%%%%%%%%%%%%%%%%%%%%%%%%%%%%%%%%%%%%%%%%%%%%%%%%%%%%%%%
\newpage

\end{document}